# Bridging Data Gaps in Healthcare: A Scoping Review of Transfer Learning in Biomedical Data Analysis

Siqi Li[1#], Xin Li[1#], Kunyu Yu[1], Di Miao[1], Mingcheng Zhu[1], Mengying Yan[2], Yuhe Ke[3], Danny D'Agostino[1], Yilin Ning[1], Qiming Wu[1], Ziwen Wang[1], Yuqing Shang[1], Molei Liu[4], Chuan Hong[2], Nan Liu[1,5,6]*

[1] Centre for Quantitative Medicine, Duke-NUS Medical School, Singapore, Singapore

[2] Department of Biostatistics and Bioinformatics, Duke University, Durham, NC, USA

[3] Department of Anesthesiology, Singapore General Hospital, Singapore, Singapore

[4] Department of Biostatistics, Columbia University Mailman School of Public Health, New York, NY, USA

[5] Programme in Health Services and Systems Research, Duke-NUS Medical School, Singapore, Singapore

[6] Institute of Data Science, National University of Singapore, Singapore, Singapore

[#] These authors contributed equally to this work.

* Correspondence: Nan Liu, Centre for Quantitative Medicine, Duke-NUS Medical School, 8 College Road, Singapore 169857, Singapore. Phone: +65 6601 6503. Email: liu.nan@duke-nus.edu.sg

## Abstract

Clinical and biomedical research in low-resource settings often faces significant challenges due to the need for high-quality data with sufficient sample sizes to construct effective models. These constraints hinder robust model training and prompt researchers to seek methods for leveraging existing knowledge from related studies to support new research efforts. Transfer learning (TL), a machine learning technique, emerges as a powerful solution by utilizing knowledge from pre-trained models to enhance the performance of new models, offering promise across various healthcare domains. Despite its conceptual origins in the 1990s, the application of TL in medical research has remained limited, especially beyond image analysis. In our review of TL applications in structured clinical and biomedical data, we screened 3,515 papers, with 55 meeting the inclusion criteria. Among these, only 2% (one out of 55) utilized external studies, and 7% (four out of 55) addressed scenarios involving multi-site collaborations with privacy constraints. To achieve actionable TL with structured medical data while addressing regional disparities, inequality, and privacy constraints in healthcare research, we advocate for the careful identification of appropriate source data and models, the selection of suitable TL frameworks, and the validation of TL models with proper baselines.

# INTRODUCTION

Healthcare-related inequalities are particularly pronounced in low-resource settings, where conducting research is challenging due to poor quality and limited sample sizes of available data[1]. For example, in the Pan-Asian Resuscitation Outcomes Study (PAROS)[2] research network dataset, several key variables, such as public access defibrillation and post-resuscitation care hypothermia, are more commonly available in countries with established emergency medical services like Japan and South Korea, but less so in Thailand, Malaysia, or Turkey[2]. These disparities in data availability lead to challenges in robust modeling, as traditional solutions that rely on straightforward adoption of external models may not be applicable. There is a growing argument that researchers should prioritize recurring local validation of artificial intelligence (AI) models over external validation[3]. With the rapid emergence of AI-based solutions, researchers are exploring ways to enhance the quality of local model development and validation by borrowing and adapting external information to support research and applications in low-resource settings. Efficiently incorporating and transferring existing information into new studies is crucial for adequately addressing local needs and conditions.

Transfer learning (TL) is a learning paradigm that leverages knowledge from a related source domain (Figure 1, Box 1) to enhance learning or modeling performance in a target domain (Figure 1, Box 1)[4]. For instance, transferring the knowledge of a resuscitation outcome prediction model involves using the PAROS dataset from country A as the source domain and country B as the target domain. In this example, the study designs (outcome, data collection process, target populations, etc.) of the source and target domains are comparable (Box 1) since they both focus on the same outcome of return of spontaneous circulation and target populations who can experience out-of-hospital cardiac arrest. In other situations, the study designs of source and target domains can differ significantly, such as data collected for different diseases[5,6], treatments[7], biological samples[8,9], and patient groups[10,11]. For example, in cancer drug sensitivity prediction, the source domain could be data from one group of patients diagnosed with cancer C, and the target domain could be data from another group of patients diagnosed with cancer D. TL is particularly beneficial in low-resource settings, where using external information can help overcome challenges posed by factors such as physician shortages and limited infrastructure resources[12]. However, the technology is also beneficial in other research scenarios, improving the quality and efficiency of studies in various healthcare settings by utilizing external knowledge from related studies.

The concept of TL was first introduced in 1995[13] to address the need for machine learning (ML) algorithms that can retain and reuse previously learned knowledge[14]. Despite its early emergence, TL has yet to fully realize its potential in analyzing structured healthcare data[15], such as electronic health records (EHRs) and data from traditional cohort studies[16]. Existing reviews on TL applications, whether in general domains or healthcare, primarily focus on methods developed by computer scientists, neglecting those from statistical domains[14,15,17–19]. This oversight has led to the underutilization of many statistical TL approaches, resulting in limitations for studies requiring more rigorous statistical analysis. A thorough examination of a broader range of TL methods could provide valuable insights for their future integration and application in the field. Additionally, the current literature[15,19] lacks comprehensive discussions on the diverse scenarios of TL applications in healthcare, particularly in cross-institutional collaborations and scenarios complicated by data privacy constraints.

The aim of this review is to (1) analyze how TL can be applied to various types of clinical and biomedical research using structured data, (2) spotlight important TL techniques, including those that have been previously overlooked or under-analyzed in the healthcare domain, and (3) provide suggestions for improving future research practices.

---

**Box 1** Transfer learning (TL) glossary

- **Source data/domain:** the original dataset used to train a model before it is applied to new data.
- **Target data/domain:** the new dataset of interest used to train new models.
- **Source model:** the original model trained using the source data/domain.
- **Target model:** the new model trained using only the target data/domain.
- **TL model:** the model trained using TL techniques.
- **Domain similarity:** the degree of similarity between the study designs of source and target data. Examples include:

---

- o   Comparable: source and target domains are both patient vital signs collected at emergency departments.
- o   Different diseases: source and target domains are both patient medical records, but one pertains to patients diagnosed with periodontal disease while the other pertains to patients diagnosed with rheumatoid arthritis.
- o   Different diseases and treatments: source and target domains are both drug sensitivity data, but one is for cisplatin in multiple myeloma patients, the other is for docetaxel in breast cancer patients.
- o   Different types of samples: source and target domains are both gene expression data but collected from different tissues.
- o   Different settings: source and target domains are both intensive care unit (ICU) data but measured with different ICU monitoring systems.
- o   Different patients: source and target domains are both EHR surgical data but involve patients with different surgical complications.
- **TL types:**
  - o   Parameter transfer: leverage parameters from the source model to update and fine-tune the target model using target data.
  - o   Feature representation: embed the good feature representations from source model to the target.
  - o   Instance re-weighting: compare the similarity between instances from source and target data and minimize model loss through weighted instances.
- **Source-free TL:** the process of TL where only the source model is used, without the source data.
- **Temporal adaptation:** the process of TL where knowledge transfer is conducted over time.

## RESULTS

Our search yielded 3,515 articles. After removing 1,775 duplicate records, 1,740 articles were screened based on their titles and abstracts. Of these, 161 articles were selected for full-text screening, and 55 articles were ultimately included in our review. Figure 1 presents the PRISMA[20] flow diagram, detailing the selection process. Notably, 47 out of the 55 reviewed papers were published in 2020 or later, indicating a recent surge of interest in TL for clinical research with structured data. As shown in Table 1, cancer medicine is the most investigated field for TL techniques, with a total of 20 papers. Additionally, in approximately 50% (27 out of 55) of the papers, the outcome types for both the source and target models are binary. In Figure 2, we present a Sankey diagram that maps the origins of both source and target data in each study. The diagram highlights that most knowledge transfer occurs between source and target data originating from developed regions, rather than involving regions with fewer research resources.

**Data and study characteristics**

We classify the articles based on their clinical types of TL studies from three perspectives, as shown in Figure 3. First, domain similarity (Box 1) includes six different scenarios observed in the reviewed studies: comparable[21,22], different diseases[5,6], different diseases and treatments[7], different types of samples[8,9], different patient groups[10,11] and different settings[23,24]. Second, the usage of source data considers whether the TL process involves only the usage of the source model without using source data or if source data is used. When source data is used, we examine whether the access to source data is restricted due to privacy constrains, such as when the source data is kept by owners who cannot share it with third parties. Third, temporal adaptation (Box 1) considers whether the TL between source and target data involves models that evolve over time.

Notably, 85% (47 out of 55) of the included papers conduct TL using source data with unrestricted access, while only one study[25] utilizes a source model obtained from an external study without using any source data. Additionally, 51% (28 out of 55) of TL instances in the included papers occur between source and target domains that are comparable. Only 5% (3 out of 55) of the reviewed articles involve TL with temporal adaptations.

**Modeling**

In this study, we categorize modeling into two types: prediction and non-prediction, based on whether the primary goal is predicting outcomes or non-prediction tasks such as investigating the relationship between covariates and outcomes or conducting phenotyping tasks[16]. For example, the study by Edmonson et al. is considered as an association study, since it focused on investigating the association between length of stay in COVID-19 patients with patient characteristics[26]. As shown in Table 1, all articles included in the review involve prediction tasks, with only four of them additionally conducting non-prediction tasks. We also summarized the types of models used by these articles, with 22 of the 55 papers involving neural networks (NNs) and 14 utilizing traditional statistical models. A total of 12 studies conducted modeling using more than one type of model.

**TL techniques**

We classified TL methods into three categories based on specific knowledge transfer techniques: parameter transfer, feature representation and instance re-weighting. As detailed in Figure 4, parameter transfer (Box 1) involves using parameters from the source model to update and fine-tune the target model using target data. Feature representation (Box 1) embeds the feature representation from the source model to the new model. Instance re-weighting (Box 1) compares the similarity between instances from source and target data, minimizing model loss through weighted instances. Among the 55 studies reviewed, 37 (67.3%) employed parameter transfer, 7 (12.7%) used feature representation, 7 (12.7%) focused on instance re-weighting, and 4 (7.3%) applied more than one type of TL strategy.

We also summarize TL methods in this review regarding their ability to preserve privacy. Source-free[27] TL (Box 1) uses only source models without any source data, making it privacy-preserving. Most parameter transfer TL methods can preserve privacy, while most instance re-weighting TL methods cannot. Detailed information on the included papers is provided in the full information extraction tables in eTables 2-4 in the Supplementary Material. As summarized in Table 1, 35 out of 55 included papers utilized privacy-preserving TL methods. Additionally, we evaluate the capability of TL methods to handle heterogeneity in covariate effects (see more details in the Discussion section), with 8 out of the 55 included papers employing such methods.

# DISCUSSION

While TL has been widely researched in the general AI field, leading to successful applications in text classification, image processing and audio analysis across various domains such as transportation and recommender systems[17,28,29], its application in healthcare research using structured data remains underexplored. Our review found that existing research has not fully tapped into TL's potential to assist in low-resource settings. This is evident as many applications require access to source data, which poses a significant challenge for low-resource environments where directly utilizing models from published works without source data is often the only viable option. Additionally, we noticed that many studies utilize TL where source data is freely accessible, while applications in cross-site collaborations with privacy constraints are limited. Furthermore, many studies lack sufficient evidence demonstrating the generalizability of their TL methods to different scenarios, especially concerning the degree of domain similarity. In this section, we provide a detailed discussion on how to fully leverage TL in low-resource settings, explore additional benefits of TL for privacy-preserving multi-site collaborations, and address technical details specific to structured clinical and biomedical data.

**Leveraging TL in low resource settings and beyond**

TL can occur in various scenarios. As shown in Figure 3, most existing TL applications have been conducted between source and target domains with comparable study designs. However, several gaps remain in other scenarios, indicating untapped potential beyond the scope of the studies reviewed. Additionally, while TL with temporal adaptation shows promise, its application is still limited and warrants further investigation.

One key advantage of TL is its ability to leverage existing knowledge from published works, as demonstrated by Hwang et al.'s study[25], the only example among the 55 papers reviewed. This application allows data owners to create new models using publicly available resources, which is especially promising in scenarios where fostering collaborations is challenging and time-consuming for less established researchers. By using source models from published works, researchers can circumvent issues associated with data sharing and collaboration when public source data is unavailable. Consequently, TL has the potential to bridge regional disparities in resource-limited settings.

**Privacy-preserving with source-free TL**

Building on the advantages of TL for low-resource settings, its role in privacy-preserving scenarios of research collaborations is another critical aspect. With the rise of cross-site collaborations in healthcare research, privacy preservation has become increasingly significant due to data-sharing constraints. Federated learning (FL) has emerged as a key technique for collaborative model training without data sharing[30,31]. Unlike FL, source-free TL can utilize existing information—such as model parameters from published works—to update individual models without a co-training process. This approach streamlines model training and also allows the incorporation of external information into new studies without re-analyzing source data. Therefore, source-free TL frameworks can achieve functions similar to FL and can be integrated with FL to further leverage federated models for client-specific model personalization and calibration.

**Heterogeneity between source and target domains**

Having discussed the potential of TL for low-resource setting and multi-site collaboration, we now delve into the capability of TL to handle varying degrees of domain similarity. Similar to the scenarios of FL[16], traditional TL techniques address data heterogeneity, broadly defined as scenarios where data are not independently and identically distributed (i.i.d.)[21,32]. In statistical literature, heterogeneity between source and target domains is usually categorized into heterogeneity in covariate effects (the conditional distribution of $Y|X$) and heterogeneity in covariate distributions ($P(X)$). While most TL methods can handle heterogenous covariate distributions, their performance in the presence of covariate effects heterogeneity has not been thoroughly examined.

TL methods such as Trans-Lasso[33], Trans-Cox[22] and COMMUTE[34], typically conceptualize a calibration term to address covariate effects heterogeneity. In this review, 8 out of 55 papers have employed such methods, demonstrating their capability to handle covariate effects heterogeneity primarily through simulation studies. Understanding how TL methods perform under different levels of domain similarity is crucial. For TL methods evaluated using limited datasets, confidence in their generalizability to different TL and research settings is lower. Researchers may consider benchmarking various TL frameworks across different simulation scenarios and real-world data examples to provide a clearer understanding of their robustness and applicability.

**Suggested steps for TL applications**

We now introduce our own suggestions for future research intending to apply TL for structured clinical and biomedical data. A summary of the ensuing discussion is outlined in Figure 5, which further visualizes these processes in detail.

| |
|---|
| **Box 2** Summary of suggestions for future research |
| - **Step 1: Identify available and appropriate source domain(s)**<br>Users should first choose candidate sources based on domain research knowledge and then decide whether or not source data will be used in the study. If source data will be used, it is crucial to determine whether access to this data is restricted due to privacy constraints. Researchers should explore pre-existing studies that provide comprehensive details about models suitable for TL adaptation as source models and potential source data, including public datasets or collaborator-owned data.<br>- **Step 2: Use TL framework that suits the research question**<br>The choice of TL methods is decided by the type of models the users want to employ. In general, the types of models can be classified into three categories: statistical models, neural networks, and others. Consider whether the intended TL scenario involves privacy constraints and whether the TL methods can handle domain dissimilarity properly.<br>- **Step 3: Validate TL models with proper baselines**<br>TL models can be evaluated against target models, source models, and pooled models depending on their availability. Ensure that TL models and target-only models do not yield conflicting outcomes based on domain knowledge. If discrepancies occur, revisit the selection of source data and TL methods and engage with experts to make appropriate adjustments. |

***Step 1: Identify available and appropriate source domain(s)***

As illustrated in Figure 5, users should first determine whether their TL process will utilize source data. If source data will not be used, the source typically originates from existing works, such as relevant publications where model parameters are provided. For instance, the TL study by Hwang et al.[25] utilized the deep neural network (DNN) model from Lee et al.[35], which was subsequently updated via TL. When using source data, users must identify whether the access to source data is restricted due to privacy constraints. If the source data is fully accessible, the TL process can proceed straightforwardly. Examples include Zhang et al.[32] using source data from their affiliated hospital, Li et al.[33] employing publicly available data, and Gu et al.[34] utilizing source data shared by collaborators without privacy constraints. When the source data is not directly accessible due to privacy constraints, users should select TL methods capable of handling such restrictions. Examples of this scenario include studies by Wang et al.[36] and Zhu et al.[37], where the source data were held by different owners and cannot be shared.

In brief, when preparing for TL, users should explore two key components: pre-existing studies that provide comprehensive details about models suitable for TL adaptation as source models and potential source data, including public datasets or collaborator-owned data, regardless of privacy constraints.

*Step 2: Use TL framework that suits the research question*

The choice of TL frameworks should be based on modeling needs and whether the intended TL scenario involves privacy constraints. Generally, most parameter transfer type TL frameworks are privacy-preserving, including those for NNs and traditional statistical regressions. On the contrary, most instance re-weighing strategies of TL, such as those used by Zhang et al.[32] and Zhu & Pu et al.[21], are not privacy-preserving since they involve calculating the distance between source and target data. It is not uncommon for TL to occur within or after multi-site collaborations, overlapping with FL scenarios, as exemplified in Wang et al[36], and in these cases, TL methods should also be privacy-preserving.

Another important consideration for the choice of TL frameworks is whether they can handle heterogeneity in covariate effects. As discussed previously, covariate effects heterogeneity is likely to exist when there are significant degrees of dissimilarity between the study designs of sources and target domains. Thus, TL methods demonstrated to handle covariate effects heterogeneity should be considered first.

*Step 3: Validate TL models with proper baselines*

Effective evaluation of TL models relies heavily on adequate baseline comparisons. As indicated in Table 1, most papers (46 out of 55) included comparisons with non-TL models. Figure 5 shows that target models and source models consistently serve as baseline options. It is crucial to ensure that TL models and target models do not yield conflicting outcomes based on domain knowledge. For instance, in prediction tasks, a significant drop in prediction performance could raise concerns. Similarly, in non-prediction tasks like association studies, a significant decrease in the confidence of statistical inference can indicate issues. If discrepancies occur, it may suggest inadequacies in the chosen TL methods or that the source knowledge is too irrelevant to the target tasks, introducing excessive noise into the study. In such cases, users should revisit the initial steps, discuss with experts familiar with the subject matter and TL techniques, and consider adjusting to the selection of source data and TL methods. Additionally, users may establish pooled models for comparisons if the source data is unrestricted. This baseline can provide further insights into the effectiveness and relevance of the TL models.

## METHODS

### Search strategy and selection criteria

This review was conducted based on the PRISMA guidelines[20]. We conducted a search for published articles that employed TL with structured clinical or biomedical data. We searched SCOPUS, MEDLINE, Web of Science, Embase, and CINAHL databases for articles published by August 22, 2023, utilizing a combination of search terms, including "electronic health records", "EHR", "electronic medical records", "EMR", "registry/registries", "tabular", "gene", "bioassay", "biological network", "transfer learning", "domain adaptation" and "knowledge transfer". A detailed search strategy is presented in eTable 1 of the Supplementary Materials.

The final search was conducted on August 22, 2023. After removing duplicates, each article was screened by at least two reviewers (S.L., X.L., K.Y., D.M., and Q.W) independently based on titles and abstracts, with a third reviewer resolving any conflicts. Publications selected in the initial screening round underwent full-text examination to ensure they met all of the following inclusion criteria: using clinical/biomedical structured data to address clinical/biomedical research questions, employing TL with provided details, being a research article, and having the full text available. We excluded studies that were irrelevant to TL, lacked clear TL details, or had unclear study designs. Additionally, studies not directly related to biomedical or clinical research were excluded. We also excluded studies that used unstructured data, were not research papers, were not peer-reviewed, or where the full text was not available.

**Data extraction**

We extracted information from the selected publications from three perspectives: study characteristics (study field, origin of reused model, participating regions, data public availability, data types, outcomes, number of participating sources and targets, number of features, sample size, study type, validation of TL model, and code availability), modeling characteristics (types of task, modeling approach, hyperparameter methods, model performance metrics), and TL characteristics (capability of privacy-preserving, types of TL, and solution for covariate effects heterogeneity).

## CONCLUSION

This review offers a comprehensive summary of TL techniques and their applications for structured medical data, categorizing them from clinical and medical perspectives. TL holds significant promise for enhancing model performance in low-resource settings by leveraging pre-existing knowledge. However, most studies only use data from developed regions, highlighting a gap in addressing resource inequalities. Additionally, privacy-preserving TL methods are underutilized. By carefully selecting source domains, employing suitable TL frameworks, and validating models against proper baselines, TL can improve healthcare outcomes and address disparities. Future research should focus on developing robust, privacy-preserving TL methods to expand their applicability in diverse healthcare contexts.

## AUTHOR CONTRIBUTIONS

**Siqi Li:** Conceptualization, Project administration, Literature search, Screening, Data extraction and analysis, Data interpretation, Writing – original draft, Writing – review & editing. **Xin Li:** Literature search, Screening, Data extraction and analysis, Writing – original draft. **Kunyu Yu:** Screening, Data extraction and analysis, Writing – original draft. **Di Miao:** Screening, Data extraction and analysis, Writing – original draft. **Mingcheng Zhu:** Data extraction and analysis, Writing – original draft. **Mengying Yan:** Data extraction and analysis, Writing – review & editing. **Yuhe Ke:** Data extraction and analysis, Data interpretation, Writing – review & editing. **Danny D'Agostino:** Data extraction and analysis, Writing – original draft. **Yilin Ning:** Writing – original draft, Writing – review & editing. **Qiming Wu:** Screening, Data extraction and analysis. **Ziwen Wang:** Writing – review & editing. **Yuqing Shang:** Writing – original draft. **Molei Liu:** Data interpretation, Writing – review & editing. **Chuan Hong:** Data interpretation, Writing – review & editing. **Nan Liu:** Conceptualization, Project administration, Funding acquisition, Resources, Supervision, Writing – review & editing.

## FUNDING

This work was supported by the Duke/Duke-NUS Collaboration grant. The funder of the study had no role in study design, data collection, data analysis, data interpretation, or writing of the report.

## COMPETING INTERESTS

All authors declare no financial or non-financial competing interests.

## SUPPLEMENTARY INFORMATION

https://drive.google.com/file/d/16RYW_8IsMEkHA-awqZD8TLo7CT2_iJSp/view?usp=sharing

**Table 1**. Summary of information extraction table.
**Figure 1**. Preferred reporting items for systematic reviews (PRISMA) flow diagram.
**Figure 2**. Sankey diagram illustrating the counts of matches between source and target regions.

**Figure 3.** Comprehensive characteristics analysis for included TL articles categorized from clinical perspectives.

**Figure 4**. Illustration of three types of TL methods.

**Figure 5**. Suggestions for future users.

**Table 1. Summary of information extraction table.**

| Model Characteristics | | No. of Papers | Examples |
|---|---|---|---|
| Task | | | |
| | Prediction | 51 | [25,32,38] |
| | Both prediction & non-prediction | 4 | [10,22,39] |
| Model type | | | |
| | Statistical Regression | | |
| | (1) Logistic regression | 3 | [40–42] |
| | (2) Lasso | 3 | [33,34,43] |
| | (3) Elastic net | 2 | [44,45] |
| | (4) Cox regression | 2 | [22,36] |
| | (5) Bayesian negative binomial model | 1 | [46] |
| | (6) Hierarchical Bayesian linear model | 1 | [47] |
| | (7) Hierarchical infinite Bayesian latent factor regression | 1 | [10] |
| | (8) High-dimensional GLM | 1 | [39] |
| | Neural networks | | |
| | (1) Multilayer perceptron | 3 | [9,25,48] |
| | (2) Recurrent neural network | 3 | [24,49,50] |
| | (3) Autoencoder | 2 | [21,51] |
| | (4) Convolutional neural network | 2 | [5,52] |
| | (5) DeepSurv | 1 | [37] |
| | (6) Generative adversarial network | 1 | [8] |
| | (7) Graph neural network | 1 | [11] |
| | (8) Transformers | 1 | [53] |
| | (9) Other | 8 | [54–56] |
| | Gradient boosting machine | 2 | [32,57] |
| | Random forest | 2 | [58,59] |
| | Gaussian graphical | 1 | [60] |
| | Gaussian process | 1 | [61] |
| | Support vector machine | 1 | [62] |
| | Multiple | 12 | [63–65] |
| **TL Characteristics** | | **No. of Papers** | **Examples** |
| Types of TL | | | |
| | Parameter transfer | 37 | [5,22,38] |
| | Instance re-weighting | 7 | [32,42,21] |
| | Feature representation | 7 | [7,8,66] |
| | Multiple | 4 | [57,63,67] |
| Capable of preserving privacy | | | |
| | Yes | 35 | [22,68,69] |
| | No | 20 | [21,32,57] |
| Strategy for heterogeneity in covariate effects | | | |
| | Yes | 8 | [34,39,60] |
| | No | 46 | [25,32,38] |

**Table 1** – continued from previous page

| Study Characteristics | | No. of Papers | Examples |
|---|---|---|---|
| Overall study field | | | |
| | Administrative/Education | 3 | [44,53,64] |
| | Cancer Medicine | 20 | [5,22,62] |
| | Cardiovascular Health | 1 | [25] |
| | Critical Care | 5 | [21,23,38] |
| | Drugs | 6 | [7,8,59] |
| | Endocrine | 1 | [34] |
| | Emergency medicine | 1 | [70] |
| | Genomics | 7 | [43,71,72] |
| | Infectious Disease | 2 | [36,48] |
| | Internal Medicine | 2 | [49,73] |
| | Neurology | 2 | [47,74] |
| | Perioperative | 1 | [10] |
| | Precision Medicine | 1 | [54] |
| | Renal | 3 | [32,40,57] |
| Outcome type (source; target) | | | |
| | (Binary; Binary) | 27 | [24,40,41] |
| | (Survival; Survival) | 8 | [21,22,38] |
| | (Continuous; Continuous) | 6 | [25,47,63] |
| | (Multi-category; Multi-category) | 3 | [54,58,65] |
| | Different pairs | 5 | [62,68,75] |
| | Not available | 6 | [23,44,60] |
| Origin of reused model | | | |
| | Current Study | 54 | [32,38,68] |
| | External source | 1 | [25] |
| Comparison with (no-TL) models | | | |
| | Yes | 46 | [25,38,57] |
| | No | 9 | [32,65,68] |
| Data public availability | | | |
| | Yes | 30 | [37,39,47] |
| | No | 25 | [23,24,44] |
| Code availability | | | |
| | Yes | 24 | [22,25,68] |
| | No | 31 | [21,32,38] |

**Figure 1. Preferred reporting items for systematic reviews (PRISMA) flow diagram.** Out of 3515 studies identified from MEDLINE, Scopus, Embase, CINAHL and Web of Science, 55 studies were included in the systematic review.

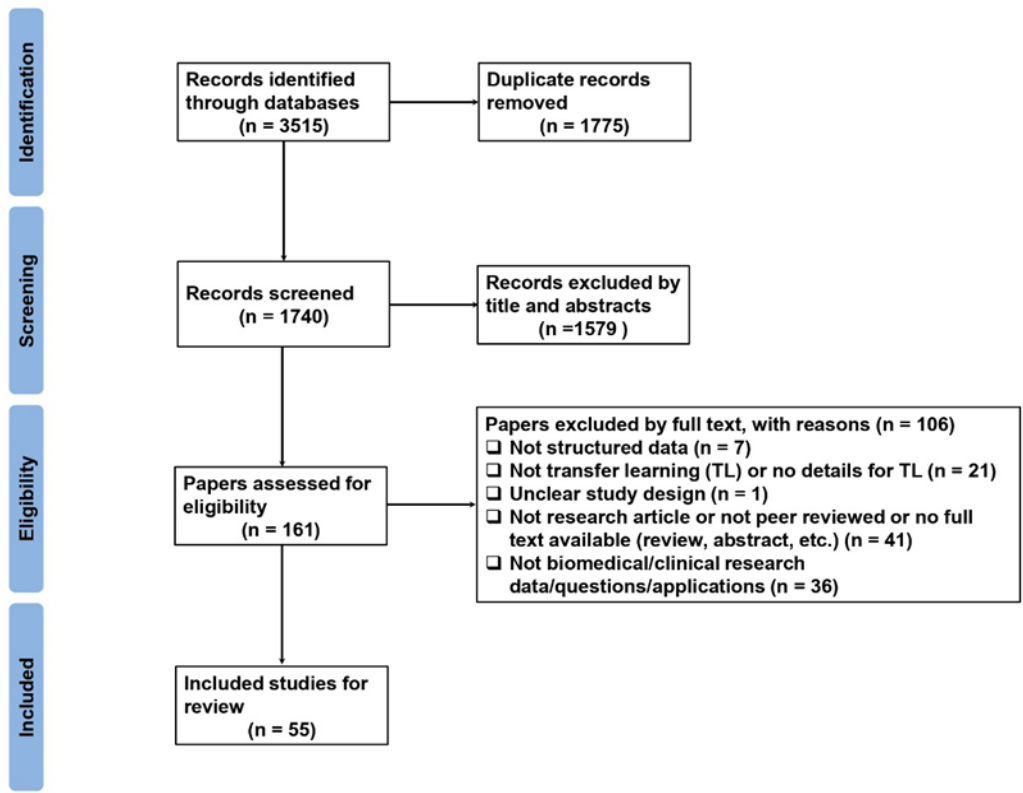

**Figure 2. Sankey diagram illustrating the counts of matches between source and target regions.** Labels indicate the number of matches between each pair. Out of 55 included papers, 49 were used for plotting. The remaining 6 papers were excluded due to indeterminable source and/or target regions or overly complex matching that did not fit into the defined categories. Note that each paper may contribute multiple matching pairs, as different sub-studies within an article can involve various source and target regions.

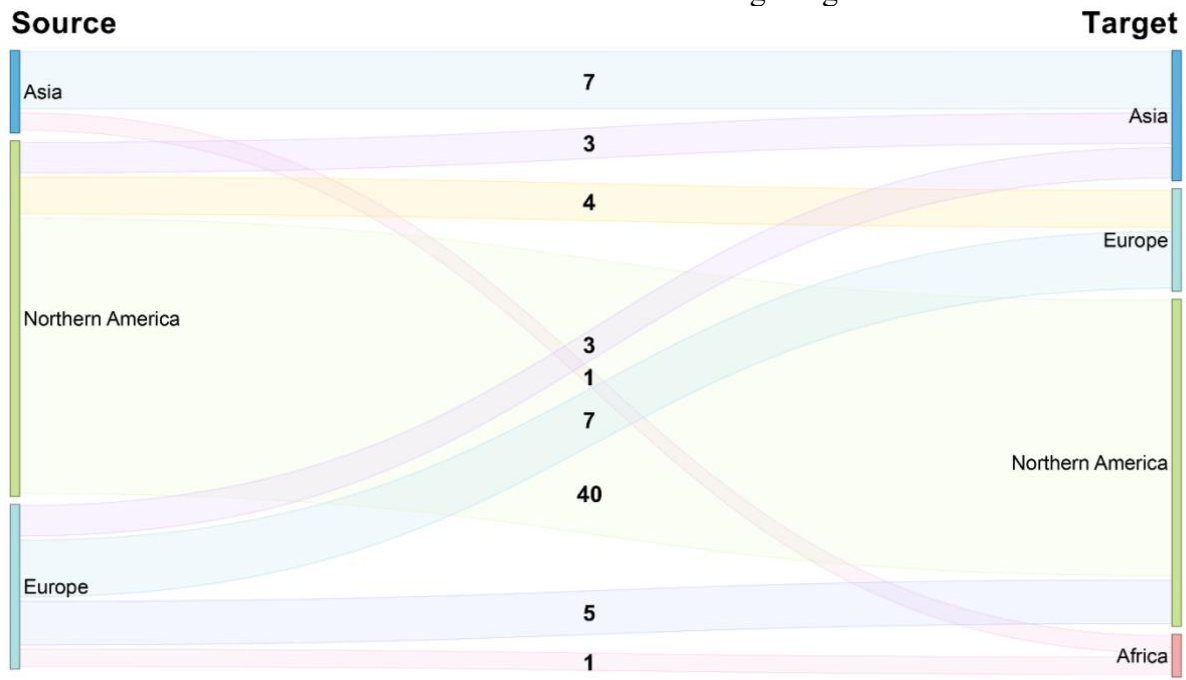

**Figure 3. Comprehensive characteristics analysis for included TL articles categorized from clinical perspectives.** Each cell indicates the number of articles that fall into a specific category, with darker shades representing a higher number of articles.

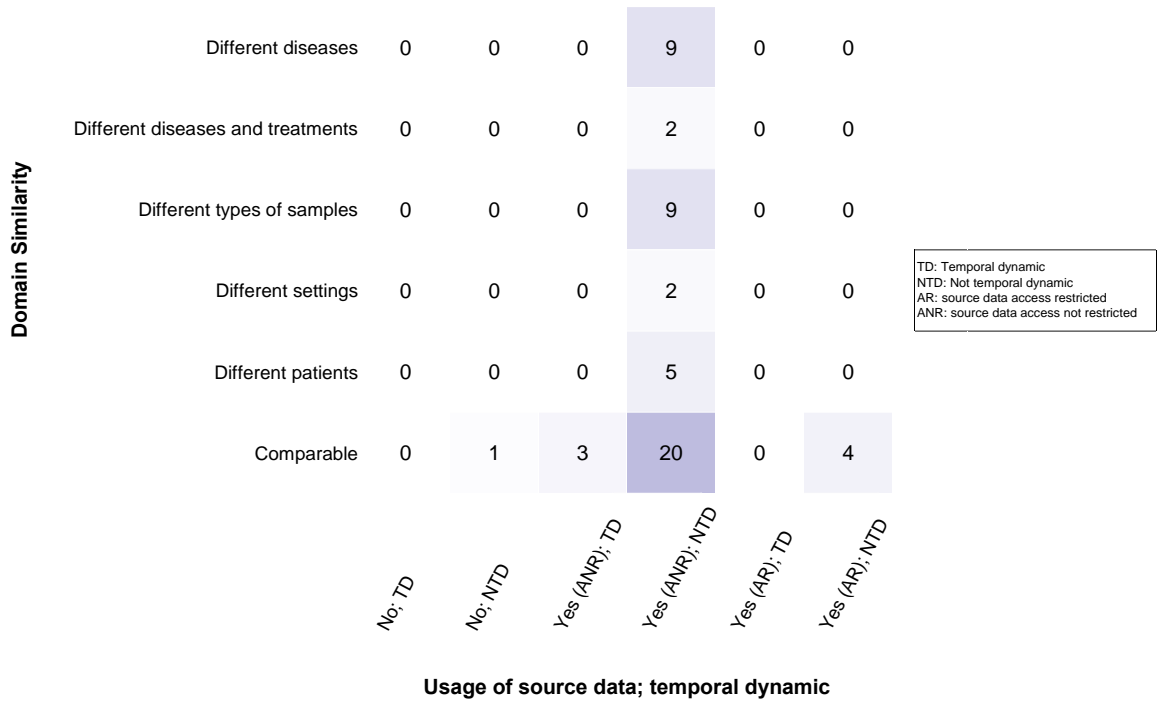

**Figure 4. Illustration of three types of TL frameworks.** (a) Parameter transfer: For neural networks, the model is pre-trained on source data and then fine-tuned with target data; for statistical models, such as Trans-Lasso[33], parameters computed on source data are calibrated using target data. (b) Feature representation: Knowledge is embedded within the learned feature representation to facilitate transfer between domains. (c) Instance re-weighting: Instances are re-weighted by comparing the similarity between instances from source data and target data to minimize model loss.

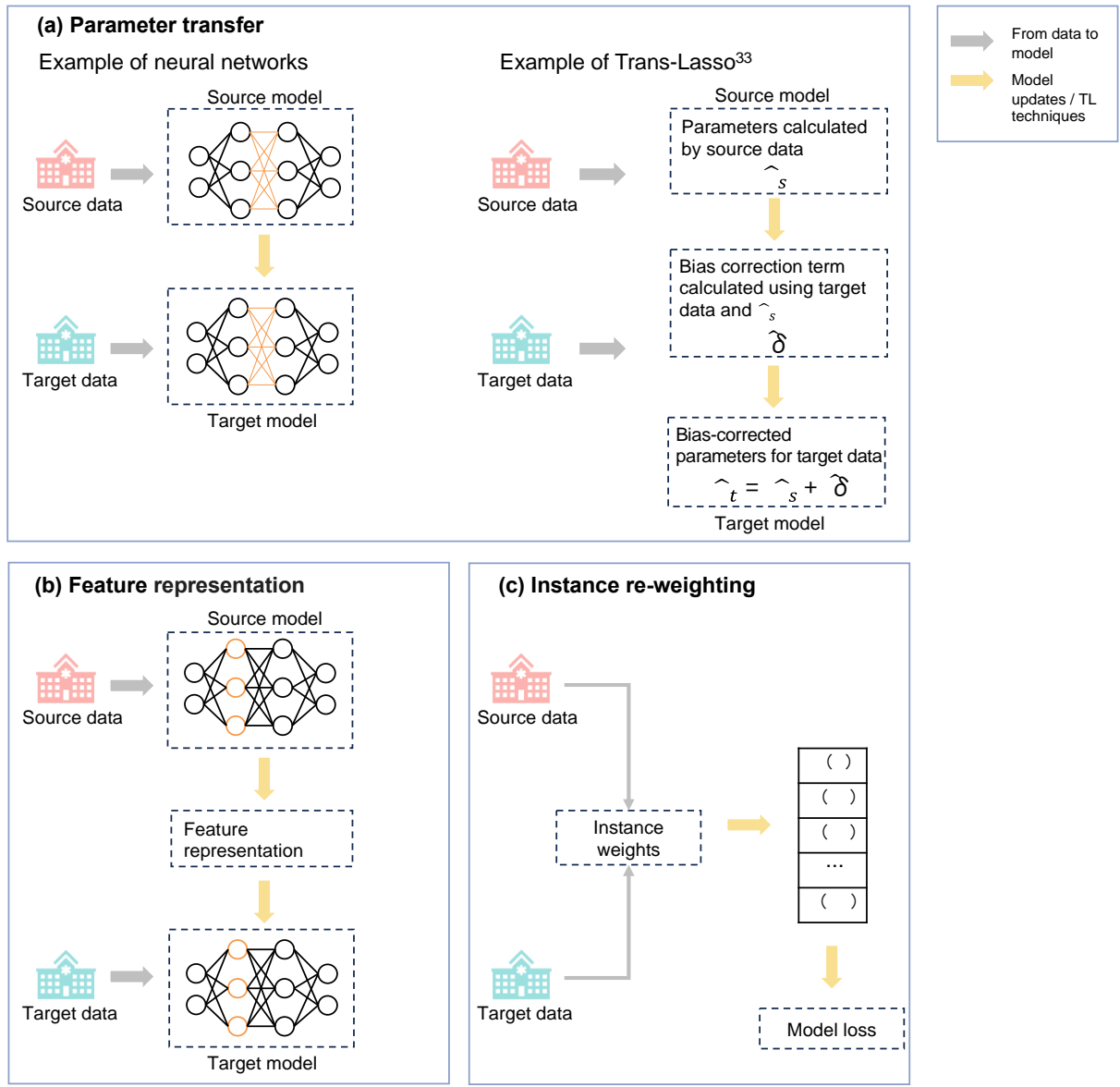

**Figure 5. Suggestions for future users.**

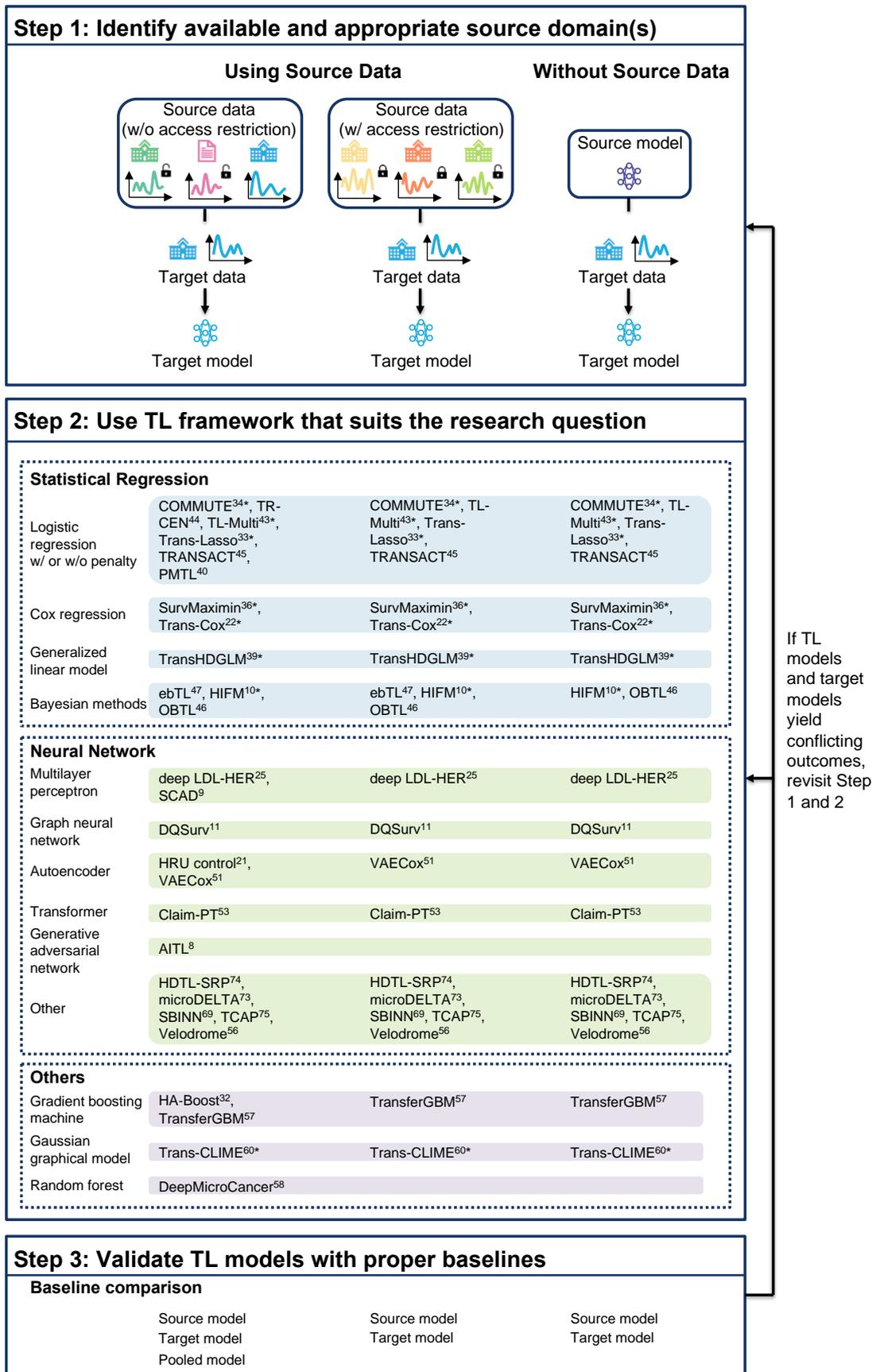